\documentclass[10pt,twocolumn,letterpaper]{article}

\usepackage{cvpr}      

\usepackage[dvipsnames]{xcolor}

\definecolor{cvprblue}{rgb}{0.21,0.49,0.74}
\usepackage[pagebackref,breaklinks,colorlinks,citecolor=cvprblue]{hyperref}

\def\paperID{11} 
\def\confName{CVPR}
\def\confYear{2024}

\title{Translating Imaging to Genomics: Leveraging Transformers for Predictive Modeling}

\author{Aiman Farooq\\
Indian Insitute of Technology Jodhpur\\
N.H. 62, Nagaur Road,\\ Karwar Jodhpur 342030\\
Rajasthan (India)\\
{\tt\small farooq.1@iitj.ac.in}
\and
Deepak Mishra\\
Indian Insitute of Technology Jodhpur\\
N.H. 62, Nagaur Road,\\ Karwar Jodhpur 342030\\
Rajasthan (India)\\
{\tt\small dmishra@iitj.ac.in}
\and
Santanu Chaudhury\\
Indian Insitute of Technology Jodhpur\\
N.H. 62, Nagaur Road,\\ Karwar Jodhpur 342030\\
Rajasthan (India)\\
{\tt\small schaudhury@gmail.com}
}

\begin{document}
\maketitle
\begin{abstract}
In this study, we present a novel approach for predicting genomic information from medical imaging modalities using a transformer-based model. We aim to bridge the gap between imaging and genomics data by leveraging transformer networks, allowing for accurate genomic profile predictions from CT/MRI images. Presently most studies rely on the use of whole slide images (WSI) for the association, which are obtained via invasive methodologies. We propose using only available CT/MRI images to predict genomic sequences. Our transformer based approach is able to efficiently generate associations between multiple sequences based on CT/MRI images alone. This work paves the way for the use of non-invasive imaging modalities for precise and personalized healthcare, allowing for a better understanding of diseases and treatment.

\end{abstract}    
\section{Introduction}
\label{sec:intro}

For diseases like cancer, imaging modalities like MRI and CT are the preferred tools for diagnosing and screening patients. However, in the context of precision oncology,  treatment planning for each patient depends upon genomic markers. Gene expressions undergo drastic changes due to cancer \cite{segal2005signatures,schmauch2020deep}. Certain specific biomarkers determine the course of action for a particular cancer; for example, mutations of the Epidermal Growth Factor Receptor (EGFR) and  Kirsten Rat Sarcoma viral oncogene (KRAS)  are associated with the occurrence of Non-Small Cell Lung Cancer (NSCLC)   \cite{skoulidis2019co}, Hypermethylation of the O(6)-Methylguanine-DNA-Methyltransferase (MGMT) gene status determines the response of a patient to radiation therapy suffering from glioblastoma \cite{rivera2010mgmt}. Genetic links can also determine the behavior and the tumor subtype. In 2021, WHO  
 proposed an improved classification criterion for central nervous system tumors, incorporating genetic information \cite{louis20212021}. \par

Therefore, the determination of gene expression and status is of paramount importance for oncology treatment selection. However, genomic testing is expensive and may not be always available. Numerous studies have been done that establish the link between genomic expressions and imaging features derived from pathological images \cite{coudray2018classification,chang2018deep}, CT \cite{ghosh2015imaging} and MRI \cite{yang2003comparing,bodalal2019radiogenomics,buda2019association}. This allows us to predict the genomic expressions using the imaging modality alone. Deep Learning (DL) for predicting genomic expressions has also seen a considerable rise \cite{ahmad2019predictive,tavolara2021deep} with CNN-based methods outperforming all previous machine learning-based models. However, most of these studies rely on whole slide images (WSI) as the imaging modality \cite{schmauch2020deep,alsaafin2023learning} or to predict the mutation status \cite{wei2023multi,ahmad2019predictive}, not the complete RNA-Seq profile.

To the best of our knowledge, this work is the first to explore the prediction of complete genomic expression through the use of most commonly available imaging modalities like CT/MRI.

\begin{figure}[t]
    \centering
    \includegraphics[scale=0.14]{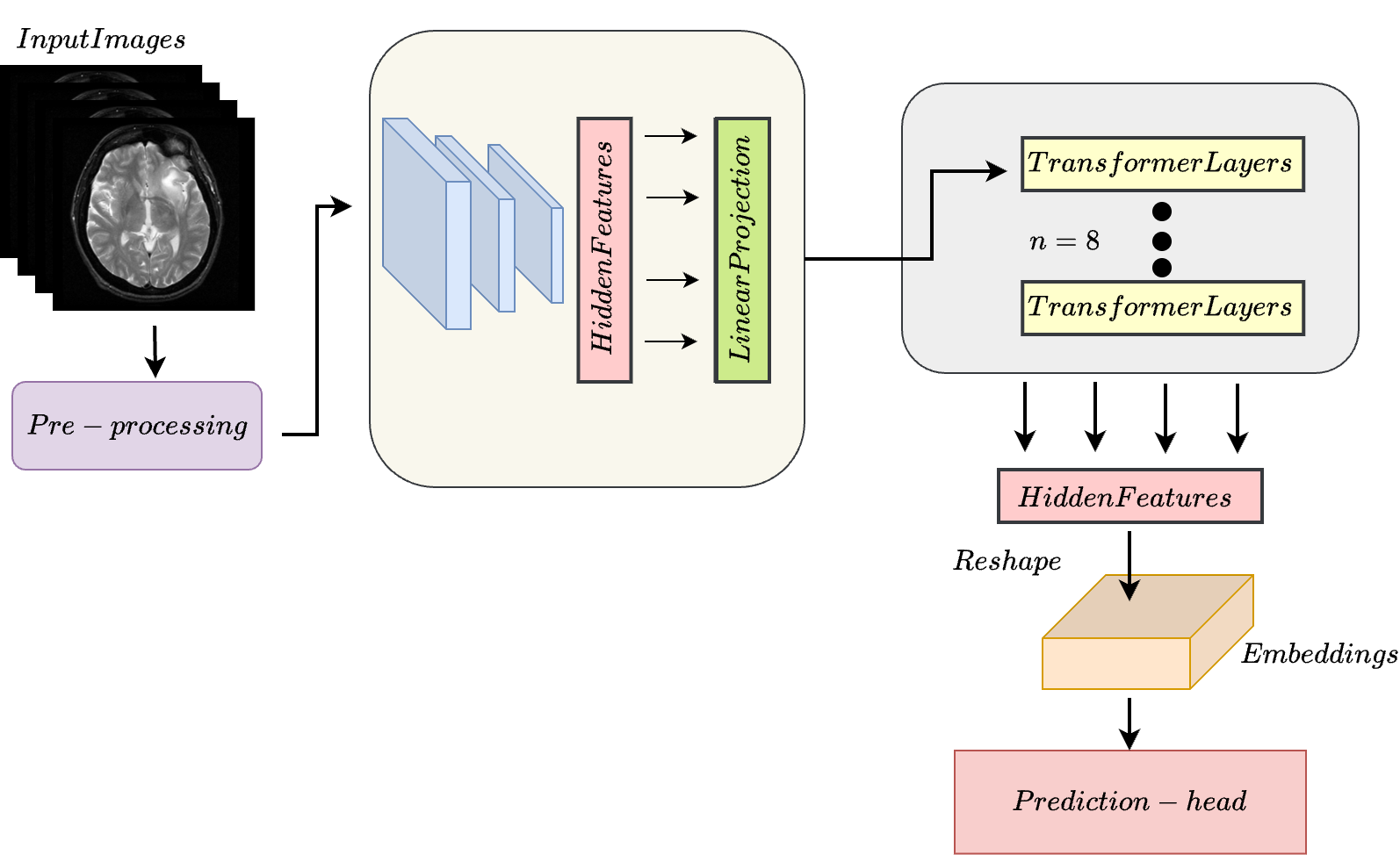}
    \caption{Overview of the workflow of the designed model. The encoder model comprises of the CNN block followed by the transformer block with 8 encoder layers. The output embeddings are passed to the gene prediction head.}
    \label{fig:over}
\end{figure}

\section{Related Work}
\label{sec:formatting}

Several studies have demonstrated the effectiveness of DL-based methods for predicting RNA-Seq from imaging data. These studies rely on pathological WSI. He \etal~\cite{he2020integrating}  used a DL-based method to predict 100 out of 30,612 gene expressions from breast cancer WSI. Zeng \etal~\cite{zeng2022spatial} proposed a Hist2ST model comprising of Convmixer, Transformer, and GNN modules and outperformed state-of-the-art models by 10\% for the HER2+ and cSCC datasets. Wang \etal~\cite{wang2021prediction} used a ResNet-based CNN for predicting gBRAC gene mutation in breast cancer patients. Saldanha \etal~\cite{saldanha2023self} integrated a self-supervised model \cite{wang2023retccl} with attention
to make patient-level predictions for a pan-cancer dataset.

All the studies above utilize WSI; researchers have also used non-invasive modalities like CT/MRI for mutation status prediction. \cite{chen2022predicting,chen2020automatic} used DL-based methods for predicting MGMT methylation status. Wei \etal~\cite{wei2023multi} proposed a multi-modal model to predict IDH mutation in glioma patients, surpassing all state-of-the-art baselines. For NSCLC lung cancer,\cite{dong2021multi,patel2021predicting} 
proposed using associations between CT and genomic mutation to predict EGFR and KRAS mutation status. Although significant efforts have been devoted to integrating RNA-Seq with whole-slide imaging (WSI), there has been limited focus on predicting complete RNA-Seq from non-invasive methods. This work entails predicting entire subsequences solely based on CT/MRI imaging modalities.


\section{Methodology}
\subsection{Datasets}
The imaging data for the study is obtained from TCIA, while the genomic data is from the TCGA portal. We have used the TCGA-GBM \cite{scarpace2016cancergenome} and TCGA-LGG \cite{albertina2016cancergenome} for brain cancer, TCGA-LUAD \cite{albertina2016cancergenome} for lung cancer and TCGA-BRCA \cite{lingle2016cancergenome} for breast cancer, respectively. We limited our study to public datasets for which CT/MRI, histopathological, and genomic data are available.

The imaging and genomic data are already anonymized, along with all the ethical clearances. For the pre-processing, CT and MRI scan slices were resampled with a slice thickness of 1 mm$^3$ and standard normalization operation was performed. Skull stripping was performed across all MRI scans. Also, we select only those scans that contain tumors. Following \cite{schmauch2020deep}, genes with a median of zero are removed, and only 31,793 genes remain. After pre-processing, we have a cohort of 550 patients with genomic, histopathological, and CT/MRI data available.

\subsection{Prediction Modelling}

The prediction module consists of a transformer encoder followed by a prediction head for genomic prediction, as shown in Fig \ref{fig:over}. We use TransUNet \cite{chen2021transunet} for the encoding process, as it captures local and global features via UNet \cite{ronneberger2015u} and the Transformer model \cite{dosovitskiy2020image}. The encoder learns the embedding for every patient. The learned embeddings are passed to the prediction head comprising of a dropout layer and 1D convolution layer similar to \cite{schmauch2020deep}. Mean Squared Error (MSE) is used as the loss function here. 
\section{Results}

\begin{figure}
    \centering
    \includegraphics[scale=0.3]{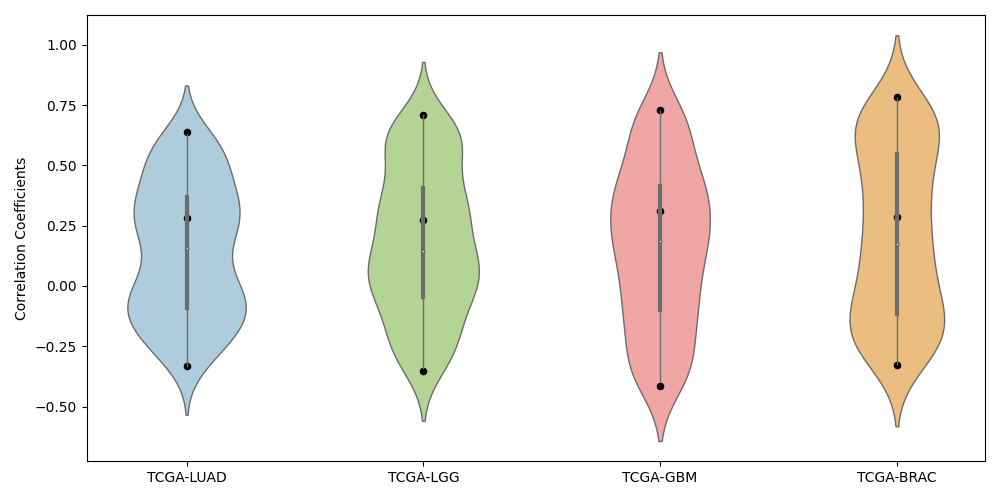}
    \caption{The distribution of Pearson correlation coefficients. The figure depicts the maximum, minimum, and mean coefficient values}
    \label{fig:pear}
\end{figure}

We evaluate our model based on two criteria similar to \cite{alsaafin2023learning,schmauch2020deep} and compare the results with He2RNA\cite{schmauch2020deep} which uses WSI. Firstly, we find Pearson correlation coefficient \cite{pearson1909determination} to determine the correlation, as shown in Fig \ref{fig:pear}. We also assess the number of genes predicted correctly using the Holm–Šidák (HS) correction discussed below.

For TCGA-GBM, we found 1731 genes with significant associations using our model, while using the He2RNA\cite{schmauch2020deep}, 3241 genes have strong correlations with WSI features. For TCGA-LGG, we found 1632 genes with significant associations, while the HE2RNA model reported 3232 genes. TCGA-BRCA 838 genes are predicted for the MRI with a statistically significant correlation; simultaneously, an average of 1627 genes are predicted with a statistically significant correlation using WSI features. For TCGA-LUAD, we saw 656 genes have significant associations, while the WSI-based model has 1539 genes with significant associations. Regarding the status of genes associated with particular forms of cancer, we found strong associations for BRAF, ALK, and KRAS for lung cancer, IDH for glioblastoma, and CHEK for breast cancer. 

The above results present strong proof of the association between imaging modalities such as CT/MRI and RNA-Seq and are encouraging enough to warrant further exploration and refinement.

\section{Conclusion}

In this study, we present a simple method to leverage transformers to establish a relationship between non-invasive modalities like CT and MRI and molecular characteristics such as RNA-Seq. The results show strong associations between multiple gene sequences and imaging features, suggesting the need for deeper exploration in this domain.
{
    \small
    \bibliographystyle{ieeenat_fullname}
    \bibliography{main}

\begin{thebibliography}{31}
\providecommand{\natexlab}[1]{#1}
\providecommand{\url}[1]{\texttt{#1}}
\expandafter\ifx\csname urlstyle\endcsname\relax
  \providecommand{\doi}[1]{doi: #1}\else
  \providecommand{\doi}{doi: \begingroup \urlstyle{rm}\Url}\fi

\bibitem[Ahmad et~al.(2019)Ahmad, Sarkar, Shah, Gore, Santosh, Saini, and Ingalhalikar]{ahmad2019predictive}
Adnan Ahmad, Srinjay Sarkar, Apurva Shah, Sonal Gore, Vani Santosh, Jitender Saini, and Madhura Ingalhalikar.
\newblock Predictive and discriminative localization of idh genotype in high grade gliomas using deep convolutional neural nets.
\newblock In \emph{2019 IEEE 16th International Symposium on Biomedical Imaging (ISBI 2019)}, pages 372--375. IEEE, 2019.

\bibitem[Albertina et~al.(2016)Albertina, Watson, Holback, Jarosz, Kirk, Lee, Rieger-Christ, and Lemmerman]{albertina2016cancergenome}
Brittany Albertina, Mark Watson, Christy Holback, Rachel Jarosz, Samantha Kirk, Yuna Lee, Kimberly Rieger-Christ, and Jordan Lemmerman.
\newblock {The Cancer Genome Atlas Lung Adenocarcinoma Collection (TCGA-LUAD) (Version 4)}.
\newblock The Cancer Imaging Archive, 2016.
\newblock Available from: \url{https://doi.org/10.7937/K9/TCIA.2016.JGNIHEP5}.

\bibitem[Alsaafin et~al.(2023)Alsaafin, Safarpoor, Sikaroudi, Hipp, and Tizhoosh]{alsaafin2023learning}
Areej Alsaafin, Amir Safarpoor, Milad Sikaroudi, Jason~D Hipp, and HR Tizhoosh.
\newblock Learning to predict rna sequence expressions from whole slide images with applications for search and classification.
\newblock \emph{Communications Biology}, 6\penalty0 (1):\penalty0 304, 2023.

\bibitem[Bodalal et~al.(2019)Bodalal, Trebeschi, Nguyen-Kim, Schats, and Beets-Tan]{bodalal2019radiogenomics}
Zuhir Bodalal, Stefano Trebeschi, Thi Dan~Linh Nguyen-Kim, Winnie Schats, and Regina Beets-Tan.
\newblock Radiogenomics: bridging imaging and genomics.
\newblock \emph{Abdominal radiology}, 44\penalty0 (6):\penalty0 1960--1984, 2019.

\bibitem[Buda et~al.(2019)Buda, Saha, and Mazurowski]{buda2019association}
Mateusz Buda, Ashirbani Saha, and Maciej~A Mazurowski.
\newblock Association of genomic subtypes of lower-grade gliomas with shape features automatically extracted by a deep learning algorithm.
\newblock \emph{Computers in biology and medicine}, 109:\penalty0 218--225, 2019.

\bibitem[Chang et~al.(2018)Chang, Grinband, Weinberg, Bardis, Khy, Cadena, Su, Cha, Filippi, Bota, et~al.]{chang2018deep}
Peter Chang, J Grinband, BD Weinberg, M Bardis, M Khy, G Cadena, M-Y Su, S Cha, CG Filippi, D Bota, et~al.
\newblock Deep-learning convolutional neural networks accurately classify genetic mutations in gliomas.
\newblock \emph{American Journal of Neuroradiology}, 39\penalty0 (7):\penalty0 1201--1207, 2018.

\bibitem[Chen et~al.(2021)Chen, Lu, Yu, Luo, Adeli, Wang, Lu, Yuille, and Zhou]{chen2021transunet}
Jieneng Chen, Yongyi Lu, Qihang Yu, Xiangde Luo, Ehsan Adeli, Yan Wang, Le Lu, Alan~L Yuille, and Yuyin Zhou.
\newblock Transunet: Transformers make strong encoders for medical image segmentation.
\newblock \emph{arXiv preprint arXiv:2102.04306}, 2021.

\bibitem[Chen et~al.(2022)Chen, Xu, Ye, Li, Sun, Liang, Lu, Wang, Zhu, Zhang, et~al.]{chen2022predicting}
Sixuan Chen, Yue Xu, Meiping Ye, Yang Li, Yu Sun, Jiawei Liang, Jiaming Lu, Zhengge Wang, Zhengyang Zhu, Xin Zhang, et~al.
\newblock Predicting mgmt promoter methylation in diffuse gliomas using deep learning with radiomics.
\newblock \emph{Journal of clinical medicine}, 11\penalty0 (12):\penalty0 3445, 2022.

\bibitem[Chen et~al.(2020)Chen, Zeng, Tong, Zhang, Fu, Li, Zhang, Cheng, Xu, Yang, et~al.]{chen2020automatic}
Xin Chen, Min Zeng, Yichen Tong, Tianjing Zhang, Yan Fu, Haixia Li, Zhongping Zhang, Zixuan Cheng, Xiangdong Xu, Ruimeng Yang, et~al.
\newblock Automatic prediction of mgmt status in glioblastoma via deep learning-based mr image analysis.
\newblock \emph{BioMed research international}, 2020, 2020.

\bibitem[Coudray et~al.(2018)Coudray, Ocampo, Sakellaropoulos, Narula, Snuderl, Feny{\"o}, Moreira, Razavian, and Tsirigos]{coudray2018classification}
Nicolas Coudray, Paolo~Santiago Ocampo, Theodore Sakellaropoulos, Navneet Narula, Matija Snuderl, David Feny{\"o}, Andre~L Moreira, Narges Razavian, and Aristotelis Tsirigos.
\newblock Classification and mutation prediction from non--small cell lung cancer histopathology images using deep learning.
\newblock \emph{Nature medicine}, 24\penalty0 (10):\penalty0 1559--1567, 2018.

\bibitem[Dong et~al.(2021)Dong, Hou, Yang, Han, Wang, Qiang, Zhao, Hou, Song, Ma, et~al.]{dong2021multi}
Yunyun Dong, Lina Hou, Wenkai Yang, Jiahao Han, Jiawen Wang, Yan Qiang, Juanjuan Zhao, Jiaxin Hou, Kai Song, Yulan Ma, et~al.
\newblock Multi-channel multi-task deep learning for predicting egfr and kras mutations of non-small cell lung cancer on ct images.
\newblock \emph{Quantitative imaging in medicine and surgery}, 11\penalty0 (6):\penalty0 2354, 2021.

\bibitem[Dosovitskiy et~al.(2020)Dosovitskiy, Beyer, Kolesnikov, Weissenborn, Zhai, Unterthiner, Dehghani, Minderer, Heigold, Gelly, et~al.]{dosovitskiy2020image}
Alexey Dosovitskiy, Lucas Beyer, Alexander Kolesnikov, Dirk Weissenborn, Xiaohua Zhai, Thomas Unterthiner, Mostafa Dehghani, Matthias Minderer, Georg Heigold, Sylvain Gelly, et~al.
\newblock An image is worth 16x16 words: Transformers for image recognition at scale.
\newblock \emph{arXiv preprint arXiv:2010.11929}, 2020.

\bibitem[Ghosh et~al.(2015)Ghosh, Tamboli, Vikram, and Rao]{ghosh2015imaging}
Payel Ghosh, Pheroze Tamboli, Raghu Vikram, and Arvind Rao.
\newblock Imaging-genomic pipeline for identifying gene mutations using three-dimensional intra-tumor heterogeneity features.
\newblock \emph{Journal of Medical Imaging}, 2\penalty0 (4):\penalty0 041009--041009, 2015.

\bibitem[He et~al.(2020)He, Bergenstr{\aa}hle, Stenbeck, Abid, Andersson, Borg, Maaskola, Lundeberg, and Zou]{he2020integrating}
Bryan He, Ludvig Bergenstr{\aa}hle, Linnea Stenbeck, Abubakar Abid, Alma Andersson, {\AA}ke Borg, Jonas Maaskola, Joakim Lundeberg, and James Zou.
\newblock Integrating spatial gene expression and breast tumour morphology via deep learning.
\newblock \emph{Nature biomedical engineering}, 4\penalty0 (8):\penalty0 827--834, 2020.

\bibitem[Lingle et~al.(2016)Lingle, Erickson, Zuley, Jarosz, Bonaccio, Filippini, Net, Levi, Morris, Figler, Elnajjar, Kirk, Lee, Giger, and Gruszauskas]{lingle2016cancergenome}
William Lingle, Bradley~J. Erickson, Margarita~L. Zuley, Rachel Jarosz, Erika Bonaccio, Jessica Filippini, James~M. Net, Lawrence Levi, Elizabeth~A. Morris, George~G. Figler, Paul Elnajjar, Samantha Kirk, Yuna Lee, Maryellen Giger, and Nicholas Gruszauskas.
\newblock {The Cancer Genome Atlas Breast Invasive Carcinoma Collection (TCGA-BRCA) (Version 3)}.
\newblock The Cancer Imaging Archive, 2016.
\newblock Available from: \url{https://doi.org/10.7937/K9/TCIA.2016.AB2NAZRP}.

\bibitem[Louis et~al.(2021)Louis, Perry, Wesseling, Brat, Cree, Figarella-Branger, Hawkins, Ng, Pfister, Reifenberger, et~al.]{louis20212021}
David~N Louis, Arie Perry, Pieter Wesseling, Daniel~J Brat, Ian~A Cree, Dominique Figarella-Branger, Cynthia Hawkins, HK Ng, Stefan~M Pfister, Guido Reifenberger, et~al.
\newblock The 2021 who classification of tumors of the central nervous system: a summary.
\newblock \emph{Neuro-oncology}, 23\penalty0 (8):\penalty0 1231--1251, 2021.

\bibitem[Patel et~al.(2021)Patel, Cowan, and Prasanna]{patel2021predicting}
Divek Patel, Connor Cowan, and Prateek Prasanna.
\newblock Predicting mutation status and recurrence free survival in non-small cell lung cancer: A hierarchical ct radiomics--deep learning approach.
\newblock In \emph{2021 IEEE 18th International Symposium on Biomedical Imaging (ISBI)}, pages 882--885. IEEE, 2021.

\bibitem[Pearson(1909)]{pearson1909determination}
Karl Pearson.
\newblock Determination of the coefficient of correlation.
\newblock \emph{Science}, 30\penalty0 (761):\penalty0 23--25, 1909.

\bibitem[Rivera et~al.(2010)Rivera, Pelloski, Gilbert, Colman, De~La~Cruz, Sulman, Bekele, and Aldape]{rivera2010mgmt}
Andreana~L Rivera, Christopher~E Pelloski, Mark~R Gilbert, Howard Colman, Clarissa De~La~Cruz, Erik~P Sulman, B~Nebiyou Bekele, and Kenneth~D Aldape.
\newblock Mgmt promoter methylation is predictive of response to radiotherapy and prognostic in the absence of adjuvant alkylating chemotherapy for glioblastoma.
\newblock \emph{Neuro-oncology}, 12\penalty0 (2):\penalty0 116--121, 2010.

\bibitem[Ronneberger et~al.(2015)Ronneberger, Fischer, and Brox]{ronneberger2015u}
Olaf Ronneberger, Philipp Fischer, and Thomas Brox.
\newblock U-net: Convolutional networks for biomedical image segmentation.
\newblock In \emph{Medical image computing and computer-assisted intervention--MICCAI 2015: 18th international conference, Munich, Germany, October 5-9, 2015, proceedings, part III 18}, pages 234--241. Springer, 2015.

\bibitem[Saldanha et~al.(2023)Saldanha, Loeffler, Niehues, van Treeck, Seraphin, Hewitt, Cifci, Veldhuizen, Ramesh, Pearson, et~al.]{saldanha2023self}
Oliver~Lester Saldanha, Chiara~ML Loeffler, Jan~Moritz Niehues, Marko van Treeck, Tobias~P Seraphin, Katherine~Jane Hewitt, Didem Cifci, Gregory~Patrick Veldhuizen, Siddhi Ramesh, Alexander~T Pearson, et~al.
\newblock Self-supervised attention-based deep learning for pan-cancer mutation prediction from histopathology.
\newblock \emph{NPJ Precision Oncology}, 7\penalty0 (1):\penalty0 35, 2023.

\bibitem[Scarpace et~al.(2016)Scarpace, Mikkelsen, Cha, Rao, Tekchandani, Gutman, Saltz, Erickson, Pedano, Flanders, Barnholtz-Sloan, Ostrom, Barboriak, and Pierce]{scarpace2016cancergenome}
Lisa Scarpace, Tom Mikkelsen, Soonmee Cha, Stephen Rao, Sharmistha Tekchandani, David Gutman, Joel~H. Saltz, Bradley~J. Erickson, Nathaniel Pedano, Adam~E. Flanders, Jill Barnholtz-Sloan, Quinn Ostrom, Daniel Barboriak, and Lori~J. Pierce.
\newblock {The Cancer Genome Atlas Glioblastoma Multiforme Collection (TCGA-GBM) (Version 5)}.
\newblock The Cancer Imaging Archive, 2016.
\newblock Available from: \url{https://doi.org/10.7937/K9/TCIA.2016.RNYFUYE9}.

\bibitem[Schmauch et~al.(2020)Schmauch, Romagnoni, Pronier, Saillard, Maill{\'e}, Calderaro, Kamoun, Sefta, Toldo, Zaslavskiy, et~al.]{schmauch2020deep}
Beno{\^\i}t Schmauch, Alberto Romagnoni, Elodie Pronier, Charlie Saillard, Pascale Maill{\'e}, Julien Calderaro, Aur{\'e}lie Kamoun, Meriem Sefta, Sylvain Toldo, Mikhail Zaslavskiy, et~al.
\newblock A deep learning model to predict rna-seq expression of tumours from whole slide images.
\newblock \emph{Nature communications}, 11\penalty0 (1):\penalty0 3877, 2020.

\bibitem[Segal et~al.(2005)Segal, Friedman, Kaminski, Regev, and Koller]{segal2005signatures}
Eran Segal, Nir Friedman, Naftali Kaminski, Aviv Regev, and Daphne Koller.
\newblock From signatures to models: understanding cancer using microarrays.
\newblock \emph{Nature genetics}, 37\penalty0 (Suppl 6):\penalty0 S38--S45, 2005.

\bibitem[Skoulidis and Heymach(2019)]{skoulidis2019co}
Ferdinandos Skoulidis and John~V Heymach.
\newblock Co-occurring genomic alterations in non-small-cell lung cancer biology and therapy.
\newblock \emph{Nature Reviews Cancer}, 19\penalty0 (9):\penalty0 495--509, 2019.

\bibitem[Tavolara et~al.(2021)Tavolara, Niazi, Gower, Ginese, Beamer, and Gurcan]{tavolara2021deep}
Thomas~E Tavolara, MKK Niazi, Adam~C Gower, Melanie Ginese, Gillian Beamer, and Metin~N Gurcan.
\newblock Deep learning predicts gene expression as an intermediate data modality to identify susceptibility patterns in mycobacterium tuberculosis infected diversity outbred mice.
\newblock \emph{EBioMedicine}, 67, 2021.

\bibitem[Wang et~al.(2021)Wang, Zou, Zhang, Xie, and Zhang]{wang2021prediction}
Xiaoxiao Wang, Chong Zou, Yi Zhang, Ling Xie, and Yifen Zhang.
\newblock Prediction of brca gene mutation in breast cancer based on deep learning and histopathology images.
\newblock \emph{Frontiers in Genetics}, 12:\penalty0 661109, 2021.

\bibitem[Wang et~al.(2023)Wang, Du, Yang, Zhang, Wang, Zhang, Yang, Huang, and Han]{wang2023retccl}
Xiyue Wang, Yuexi Du, Sen Yang, Jun Zhang, Minghui Wang, Jing Zhang, Wei Yang, Junzhou Huang, and Xiao Han.
\newblock Retccl: Clustering-guided contrastive learning for whole-slide image retrieval.
\newblock \emph{Medical image analysis}, 83:\penalty0 102645, 2023.

\bibitem[Wei et~al.(2023)Wei, Chen, Zhu, Zhang, Sch{\"o}nlieb, Price, and Li]{wei2023multi}
Yiran Wei, Xi Chen, Lei Zhu, Lipei Zhang, Carola-Bibiane Sch{\"o}nlieb, Stephen Price, and Chao Li.
\newblock Multi-modal learning for predicting the genotype of glioma.
\newblock \emph{IEEE Transactions on Medical Imaging}, 2023.

\bibitem[Yang et~al.(2003)Yang, Guccione, and Bednarski]{yang2003comparing}
Yi-Shan Yang, Samira Guccione, and Mark~D Bednarski.
\newblock Comparing genomic and histologic correlations to radiographic changes in tumors: a murine scc vii model study1.
\newblock \emph{Academic radiology}, 10\penalty0 (10):\penalty0 1165--1175, 2003.

\bibitem[Zeng et~al.(2022)Zeng, Wei, Yu, Yin, Yuan, Li, Tang, Lu, and Yang]{zeng2022spatial}
Yuansong Zeng, Zhuoyi Wei, Weijiang Yu, Rui Yin, Yuchen Yuan, Bingling Li, Zhonghui Tang, Yutong Lu, and Yuedong Yang.
\newblock Spatial transcriptomics prediction from histology jointly through transformer and graph neural networks.
\newblock \emph{Briefings in Bioinformatics}, 23\penalty0 (5):\penalty0 bbac297, 2022.

\end{thebibliography}
}

\end{document}